\pdfoutput=1

\documentclass[11pt]{article}

\usepackage{algorithm}
\usepackage{algpseudocode}

\usepackage{acl}

\usepackage{times}
\usepackage{latexsym}

\usepackage[T1]{fontenc}

\usepackage[utf8]{inputenc}

\usepackage{microtype}

\usepackage{graphicx}
\usepackage{multirow}
\usepackage{makecell}
\usepackage{amsmath}
\usepackage{amsfonts}
\usepackage{amssymb}
\usepackage{xspace}

\newcommand{\modelname}{HPD\xspace}
\newcommand{\hmodelname}{\textbf{H}omomorphic \textbf{P}rojective \textbf{D}istillation\xspace}
\title{Compressing Sentence Representation for Semantic Retrieval via Homomorphic Projective Distillation}
\author{Xuandong Zhao$^{\dagger}$, ~~~ Zhiguo Yu$^\ddagger$,  ~~~ Ming Wu$^\ddagger$, ~~~  Lei Li$^\dagger$  \\
  $^\dagger$UC Santa Barbara ~~~~ $^\ddagger$Microsoft \\
  \texttt{\{xuandongzhao,leili\}@cs.ucsb.edu} \\
  \texttt{\{zhiguo.yu,mingwu\}@microsoft.com} 
}

\begin{document}
\maketitle
\begin{abstract}
How to learn highly compact yet effective sentence representation? Pre-trained language models have been effective in many NLP tasks. However, these models are often huge and produce large sentence embeddings. Moreover, there is a big performance gap between large and small models. In this paper, we propose \hmodelname(\modelname) to learn compressed sentence embeddings. Our method augments a small Transformer encoder model with learnable projection layers to produce compact representations while mimicking a large pre-trained language model to retain the sentence representation quality. We evaluate our method with different model sizes on both semantic textual similarity (STS) and semantic retrieval (SR) tasks. Experiments show that our method achieves 2.7-4.5 points performance gain on STS tasks compared with previous best representations of the same size. In SR tasks, our method improves retrieval speed (8.2$\times$) and memory usage (8.0$\times$) compared with state-of-the-art large models. Our implementation is available at \url{https://github.com/XuandongZhao/HPD}.
\end{abstract}
\section{Introduction}
It is a fundamental problem to learn compact yet effective sentence representations. Good representations have wide applications in NLP, including web search \cite{Palangi2016DeepSE}, question answering \cite{Hao2019ExploitingSE}, knowledge inference \cite{Wang2020SBERTWKAS}, and machine translation \cite{Yang2020TowardsMT}. Sentence embedding models take a sentence as the input and output a fixed-length continuous vector representation. Based on BERT \cite{Devlin2019BERTPO} and RoBERTa \cite{Liu2019RoBERTaAR}, recent sentence embedding models such as Sentence-BERT (SBERT) \cite{reimers-2019-sentence-bert} and SimCSE \cite{gao2021simcse}, are fine-tuned on sentence pair scoring tasks to learn better sentence representations, which show much improvement in downstream tasks. However, these models are big in two aspects. 1) They contain hundreds of millions to billions of parameters, which requires large memory and powerful machine to serve in production; 2) Their resulting embeddings are high dimensional (e.g. 1024), requiring huge database to store and index, which cause high search latency. Therefore, it is challenging to directly use these large models in real-world applications with strict throughput/latency requirement and bounded hardware resources. Our work focuses on reducing both the model size and the representation size.
\begin{figure}[tb]
\centering
\includegraphics[width=0.45\textwidth]{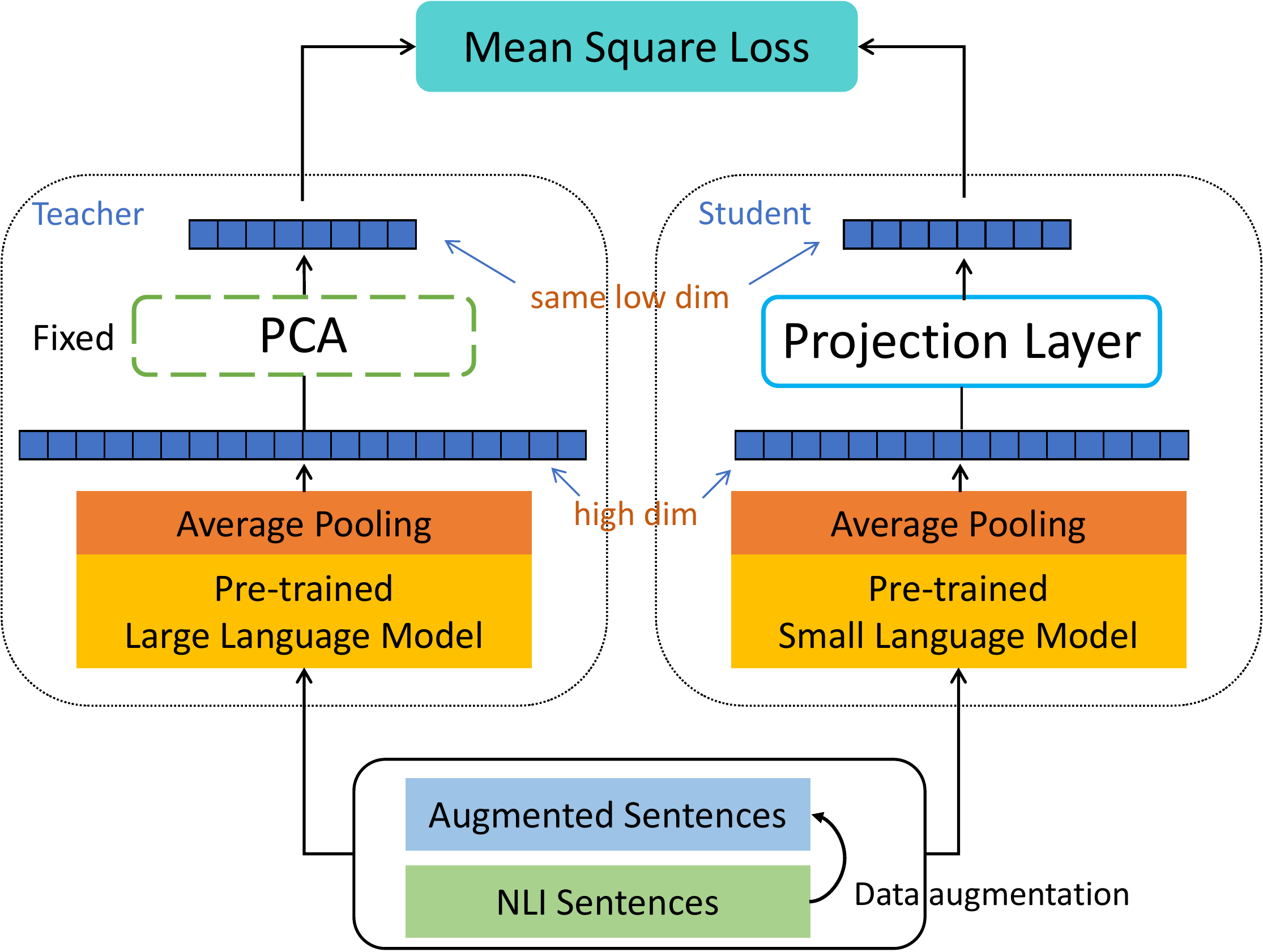}
\caption{
Overview of \textbf{H}omomorphic \textbf{P}rojective \textbf{D}istillation  (HPD). In contrast to the teacher model, which is a large pre-trained language model with a fixed PCA dimension reduction module, the student model is a smaller language model with a learnable projection layer. The mean square error of both dimension reduction results is used to train the student model.}
\label{fig:main}
\end{figure}
There has been several works to reduce model size and retain the superior model performance. Recent studies \cite{jiao-etal-2020-tinybert, Sanh2019DistilBERTAD, Wang2020MiniLMDS} have used knowledge distillation (KD) on large language models to derive compressed compact models with decent performance. TinyBERT \cite{jiao-etal-2020-tinybert} performs layer-to-layer transformer distillation at pre-training and task-specific learning stage utilizing the teacher's hidden states and self-attention distributions. MiniLM \cite{Wang2020MiniLMDS} proposes task-agnostic transformer distillation, which uses self-attention distributions and value relations to help the student deeply mimic the teacher’s self-attention modules. Nevertheless, directly fine-tuning small transformer models for sentence embedding shows less desirable results than large ones \cite{reimers-2019-sentence-bert, SentenceTransformer}.

Can we learn a compact yet highly performant sentence representation? In this work, we propose \modelname{}: a dimension reduced sentence embedding model via projected knowledge distillation. The key idea is to start from a pre-trained large model and distill its knowledge into a small one. The large model is fine-tuned on natural language inference (NLI) datasets first. Then the small and large ones are augmented with linear projection layer and Principal Component Analysis (PCA) \cite{Abdi2010PrincipalCA} respectively to reduce final representation dimension. In this way, the small model is expected to produce semantic meaningful representations (semantically similar sentences will have close embeddings), where it mimics the power of large models through homomorphic mappings.

We evaluate our model on semantic textual similarity (STS) and semantic retrieval (SR) tasks. Empirical results show that our model can attain 2.7-4.5 points of performance gain on STS tasks compared to other dimension reduction approaches and achieve competitive retrieval performance against large sentence embedding models while significantly improving retrieval speed (8.2$\times$) and memory usage (8.0$\times$) in SR tasks.



\section{Related Work}
Embedding techniques are used to represent complex data mathematically \cite{Mikolov2013EfficientEO, zhao2020multi, Khrulkov2020HyperbolicIE}. Sentence embedding is a well-researched topic with a plethora of proposed approaches. Early works \cite{Kiros2015SkipThoughtV, Logeswaran2018AnEF} build upon the distributional hypothesis and train the models to predict the surrounding sentences. Sent2Vec \cite{Pagliardini2018UnsupervisedLO} generate sentence embeddings using word vectors along with n-gram embeddings. \citet{Conneau2017SupervisedLO} propose to fine-tune a Siamese network on NLI datasets, which is then further extended to pre-trained models in Sentence-BERT \cite{reimers-2019-sentence-bert}. SimCSE \cite{gao2021simcse} proposes a contrastive learning method and achieves state-of-the-art performance on STS tasks.

Recently, \citet{Raunak2019EffectiveDR} combines PCA based dimensionality reduction with a post-processing algorithm to address the latency and capacity issues of large dimensionality models. \citet{Shi2018UnderstandingAI} proposed extended robust PCA (Ex-RPCA) to do dimension reduction. But both of them only focus on word embedding. \citet{Su2021WhiteningSR} find that the whitening operation can enhance the isotropy of sentence distribution and reduce the dimensionality of the sentence representation, which optimizes the memory storage and accelerates the retrieval speed. We use this approach as one of our baselines.

\section{Method}
The overview of our approach is illustrated in Figure \ref{fig:main}. Given a set of sentences $\mathcal{X} = \{x_i\}_{i=1}^{m}$, our goal is to obtain efficient sentence embedding models $f : \mathcal{X} \rightarrow \mathbb{R}^{d} $, where $d$ is the embedding dimension. 

The teacher model $f_t$ is trained on the same NLI dataset as \citet{Conneau2017SupervisedLO, reimers-2019-sentence-bert, gao2021simcse}, where there are three types of sentence pairs (entailment/neutral/contradiction). We follow the supervised contrastive training framework in SimCSE \cite{gao2021simcse} and take a cross-entropy objective with in-batch negatives and hard negatives. The idea is based on the assumption that a good semantic representation should be able to bring similar sentences together while pushing dissimilar ones apart. Let $(\mathbf{e}_i, \mathbf{e}_i^+, \mathbf{e}_i^-)$ denote the representations of sentence triplet $(x_i, x_i^+, x_i^-)$, where $(x_i^+, x_i^-)$ are corresponding "entailment" and "contradiction" pairs for $x_i$ in the NLI dataset. For a mini-batch with $N$ pairs, the training objective is 

\begin{equation}
\small
\ell_{i}=-\log \frac{e^{\operatorname{sim}\left(\mathbf{e}_{i}, \mathbf{e}_{i}^{+}\right) / \tau}}{\sum_{j=1}^{N}\left(e^{\operatorname{sim}\left(\mathbf{e}_{i}, \mathbf{e}_{j}^{+}\right) / \tau}+e^{\operatorname{sim}\left(\mathbf{e}_{i}, \mathbf{e}_{j}^{-}\right) / \tau}\right)},
\label{eq:1}
\end{equation}

\noindent where $\tau$ is a temperature hyperparameter; $\operatorname{sim}\left(\mathbf{e}_{1}, \mathbf{e}_{2}\right)$ is the cosine similarity $\frac{\mathbf{e}_{1}^{\top} \mathbf{e}_{2}}{\left\|\mathbf{e}_{1}\right\| \cdot\left\|\mathbf{e}_{2}\right\|}$.

After building up a teacher model, we use it for knowledge distillation. Firstly, we enrich the training dataset by data augmentation (Details in Section \ref{sec:DA}). Then for each sentence $x_i$, we get the embedding $\mathbf{e}^t_i \in \mathbb{R}^{d^\prime_t}$ from the teacher model $f_t$ and $\mathbf{e}^s_i \in \mathbb{R}^{d^\prime_s}$ from the student model $f_s$. Note that the dimensions for $\mathbf{e}^t_i$ and $\mathbf{e}^s_i$ may be different ($d^\prime_t \neq d^\prime_s$). 

\renewcommand{\algorithmicrequire}{\textbf{Input:}}
\renewcommand{\algorithmicensure}{\textbf{Output:}}
\begin{algorithm}[t]
\caption{PCA}\label{alg:cap}
\begin{algorithmic}[1]
\Require Initial embeddings $\{\mathbf{e}_i^t\}^m_{i=1}$ and reserved dimensionality $d$
\State compute $\bar{\mathbf{e}}^t$ of $\{\mathbf{e}_i^t\}^m_{i=1}$
\State compute $\mathbf{V},\mathbf{S},\mathbf{U}^\top =$ SVD$(\mathbf{E})$
\State compute $\mathbf{W}^t = \mathbf{V}[:, :d]$
\For{$i=1,2,..., m$}
    \State $\mathbf{h}_i^t = {\mathbf{W}^t}^\top(\mathbf{e}_i^t-\bar{\mathbf{e}}^t)$
\EndFor
\Ensure Transformed embeddings $\{\mathbf{h}_i^t\}^m_{i=1}$
\end{algorithmic}
\end{algorithm}

In order to perform homomorphic projective distillation, we employ Principle Component Analysis (PCA) \cite{Abdi2010PrincipalCA}, $\mathbf{h}^t_i={\mathbf{W}^t}^\top (\mathbf{e}^t_i - \Bar{\mathbf{e}}^t)$, to the teacher model after its average pooling layer and we add a projection layer, $\mathbf{h}^s_i = {\mathbf{W}^s}^\top \mathbf{e}^s_i + \mathbf{b}^s$, to the student model, where $\mathbf{h}^t_i, \mathbf{h}^s_i \in \mathbb{R}^{d}$ are the teacher and student's final embeddings with the same dimension. $\mathbf{W}^t \in \mathbb{R}^{d^\prime_t \times d}$ is the PCA weight matrix for the teacher model. $\mathbf{W}^s \in \mathbb{R}^{d^\prime_s \times d}$ is the weight matrix of the projection layer for the student model.  $\mathbf{b}^s$ is the bias term and both $d^\prime_s$ and $d^\prime_t$ are larger than the final embedding dimension $d$.

Algorithm 1 shows how to conduct the PCA over a set of initial sentence embeddings for the teacher model. We sample $m$ sentences from the training dataset and get the embeddings $\{\mathbf{e}_i^t\}_{i=1}^m$ after the average pooling layer of the teacher model. We then construct a centered matrix $\mathbf{E}$ of $d^\prime_t \times m$ size, where $d^\prime_t$ is the initial embedding dimension. Thus the $i$-th column of $\mathbf{E}$ corresponds to $\mathbf{e}_i^t-\bar{\mathbf{e}}^t$, i.e. row means have been subtracted. Then we perform the singular value decomposition (SVD) of $\mathbf{E}$ (line 2 in Algorithm 1).
Because only the first $d$ principle components are needed, we reserve the first $d$ columns of $\mathbf{V}$ (i.e. the first $d$ eigenvectors), which we denote as weight matrix $\mathbf{W}^t$. The transformed embeddings $\mathbf{h}_i^t$ in the new PC space are given by the $i$-th column of ${\mathbf{W}^t}^\top \mathbf{E}$ (line 5 in Algorithm 1).


Note that PCA only requires $m$ sample sentences and calculating their initial embeddings. So during the distillation process, the teacher's transformer parameters $\theta_t$ and PCA weight matrix $\mathbf{W}^t$ are fixed, while the student's transformer parameters $\theta_s$, projection weight $\mathbf{W}^s$, and projection bias $\mathbf{b}^s$ can be updated. We minimize the distance between final embeddings $\mathbf{h}^t_i$ and $\mathbf{h}^s_i$ by taking the mean squared loss:

\begin{equation}
	\mathcal{L} = \frac{1}{M}\sum_{i=1}^{M}\left\|\mathbf{h}^s_{i}-\mathbf{h}^t_i\right\|_{2}^{2},
\label{eq:2}
\end{equation}

\noindent where $M$ is the total number of sentences after data augmentation.

\begin{table*}[t] \resizebox{\textwidth}{!}{
\small
\centering
\begin{tabular}{lccccccc|c|c|c|c}
\Xhline{2\arrayrulewidth} 
Model & STS12 & STS13 & STS14 & STS15 & STS16 & STS-B & SICK-R & Avg. & Size & Dim & Speed\\ 
\hline\hline
\multicolumn{12}{c}{Large models}\\
SBERT\textsubscript{base}						
& 70.97 & 76.53 & 73.19 & 79.09 & 74.30 & 77.03 & 72.91 & 74.89 & 109M & 768 & 993\\
SRoBERTa\textsubscript{large} 
& 74.53 & 77.00 & 73.18 & 81.85 & 76.82 & 79.10 & 74.29 & 76.68 & 355M & 1024 &385\\
SimCSE-MPNet $\spadesuit$
& 73.70 & 86.78 & 82.56 & 87.24 & 83.06 & 86.54 & 79.27 & 82.75 & 109M & 768 & 986\\
SimCSE-RoBERTa\textsubscript{large} $\clubsuit$
& 77.46 & 87.27 & 82.36 & 86.66 & 83.93 & 86.70 & 81.95 & 83.76 & 355M & 1024 & 291\\
\hline
\multicolumn{12}{c}{Backbone for compact model: TinyBERT}\\
SimCSE-TinyBERT 
& 73.02	& 80.71	& 76.89	& 83.01	& 78.57	& 81.10	& 78.19	& 78.78 & 14M & 312 & 2650\\
~~+Projection-128
& 72.73	& 79.81	& 76.60	& 82.70	& 77.37	& 80.24	& 77.41	& 78.12 & 14M & 128 & 2604\\
~~+Whitening-128
& 73.00	& 80.81	& 77.02	& 82.79	& 78.45	& 80.97	& 78.16	& 78.74 & 14M & 128 & 2612\\
\modelname{}-128 (Teacher: $\spadesuit$)
& 74.20	& \textbf{84.49} & \textbf{79.95}	& \textbf{85.79} & 80.07 & 83.41 & 78.99 & 80.99 & 14M & 128 & 2608\\
\modelname{}-128 (Teacher: $\clubsuit$)
&\textbf{74.29} &  83.05 & 78.80 & 84.62 & \textbf{81.17} & \textbf{84.36} & \textbf{80.83} & \textbf{81.02} & 14M & 128 & 2609\\
\hline
\multicolumn{12}{c}{Backbone for compact model: MiniLM}\\
SimCSE-MiniLM
& 70.34	& 78.59	& 75.08	& 81.10	& 77.74	& 79.39	& 77.85	& 77.16 & 23M &  384 & 2031\\
~~+Projection-128
& 70.19	& 79.22	& 75.53	& 80.78	& 78.13	& 79.45	& 77.46	& 77.25 & 23M &  128 & 2022\\
~~+Whitening-128
& 70.55	& 78.85	& 75.4	& 81.06	& 77.77	& 79.40	& 77.92	& 77.28 & 23M &  128 & 2015\\
\modelname{}-128 (Teacher: $\spadesuit$)
& 74.25 & 84.43	& \textbf{80.33} & \textbf{85.75} & 80.68 & 83.91 & 79.06 & 81.20 & 23M &  128 & 2025\\
\modelname{}-128 (Teacher: $\clubsuit$)
& \textbf{74.94} &  \textbf{84.52} & 80.25 & 84.87 & \textbf{81.90} & \textbf{84.98} & \textbf{81.15} & \textbf{81.80} & 23M &  128 & 2024 \\
\Xhline{2\arrayrulewidth}
\end{tabular}}
\caption{Sentence embedding performance on STS tasks (Spearman's correlation $\rho \times 100$). STS12-STS16: SemEval 2012-2016, STSb: STS benchmark, SICK-R: SICK relatedness dataset, Dim: embedding dimension, Size: number of parameters, Speed: sentences per second.}
\label{table:main}
\end{table*}

\section{Experiment}
We conduct our experiments on standard semantic textual similarity (STS) tasks using the SentEval toolkit \cite{Conneau2017SupervisedLO} for evaluation. We also test mean reciprocal rank (MRR), memory usage, and retrieval speed on semantic retrieval (SR) tasks. All of our experiments are tested on a server with Intel i7-5930K CPU @ 3.50GHz, Nvidia TITAN Xp GPU, CUDA 11.3 and cuDNN.

\subsection{Semantic Textual Similarity (STS) Task} 
Semantic textual similarity (STS) is a natural language processing (NLP) task to quantitatively assess the semantic similarity between two text snippets. We evaluate our model by computing the cosine similarity between sentence pair embeddings on 7 standard STS tasks: STS 2012–2016 \cite{Agirre2012SemEval2012T6, Agirre2013SEM2S, Agirre2014SemEval2014T1, Agirre2015SemEval2015T2, Agirre2016SemEval2016T1} , STS Benchmark \cite{Cer2017SemEval2017T1} and SICK-Relatedness \cite{Marelli2014ASC}. These datasets
have labels between 0 and 5 indicating the semantic
relatedness of sentence pairs. Following \citet{reimers-2019-sentence-bert, Su2021WhiteningSR, gao2021simcse}, we use Spearman rank correlation to measure the correlation quality between calculated similarity and human labels. Spearman correlation has a value between -1 and 1, which will be high when the ranks of predicted similarities and the ground-truth are similar.

\subsection{Semantic Retrieval (SR) Task}
The semantic retrieval (SR) task is to identify all sentences in the retrieval corpus that are semantically similar to a query sentence. We construct the SR task on Quora Duplicate Questions Dataset\footnote{https://quoradata.quora.com/First-Quora-Dataset-Release-Question-Pairs} and Faiss\footnote{https://github.com/facebookresearch/faiss} \cite{JDH17} to test the retrieval effect and efficiency of different models.  The Quora dataset contains over 500k sentences with over 400k pairwise annotations on whether two questions are duplicates or not. Faiss \cite{JDH17} is a library for efficient similarity search and clustering of dense vectors, which contains algorithms that search in sets of vectors of any size.  We calculate all the sentence embeddings of question2, store them in Faiss, and then use the sentence embedding of question1 to retrieve them. Faiss is configured in CPU mode with 'nlist = 1024' and 'nprobe = 5'. Note that we didn't fine-tune the models on the semantic retrieval task. We report the results on three parts: average mean reciprocal ranking (MRR@10), average retrieve time for 1,000 sentences (Time/ms) and memory usage (Mem/MB).

\subsection{Experiment Setup}
We train our model on the NLI dataset, which is a combination of the SNLI \cite{Bowman2015ALA} and the MNLI \cite{Williams2018ABC} dataset. SNLI dataset contains 570k sentence pairs and MNLI is a collection of 430k sentence pairs. Particularly, the teacher model is trained on "entailment" and "contradiction" pairs in NLI dataset using contrastive loss (Equation \ref{eq:1}). We use two state-of-the-art large sentence embedding models, SimCSE-RoBERTa\textsubscript{large}\footnote{https://huggingface.co/princeton-nlp/sup-simcse-roberta-large} \cite{gao2021simcse} and SimCSE-MPNet\footnote{https://huggingface.co/sentence-transformers/nli-mpnet-base-v2} \cite{Song2020MPNetMA}, as our teacher models. For the student model, we choose the released pre-trained checkpoints of TinyBERT \cite{jiao-etal-2020-tinybert} and MiniLM \cite{Wang2020MiniLMDS}, and we leverage a linear projection layer for dimension reduction. As for the PCA implementation, we first sample 100k random sentences from the dataset and pass them to the teacher model. Then we calculate the principal components $\mathbf{W}^t$ of the output embeddings by calling the \texttt{PCA} function of the \texttt{scikit-learn} package.

\paragraph{Baseline Models}We compare our \modelname{} method to both state-of-the-art sentence embedding models and various dimension reduction techniques. For sentence embedding model baseline, we directly fine-tune pre-trained language models TinyBERT/MiniLM given NLI dataset using contrastive loss. For the dimension reduction baseline, we test both projection and whitening approaches: 1) adding a projection layer after TinyBERT/MiniLM encoder and training on NLI dataset with contrastive loss (without distillation); 2) adopting whitening \cite{Su2021WhiteningSR} as a post-processing operation (similar to PCA) to reduce the dimension of SimCSE-TinyBERT or SimCSE-MiniLM. More details about each baseline and training setting can be found in Appendix \ref{sec:appendix_a}.

\paragraph{Data Augmentation}\label{sec:DA} Data Augmentation is a set of techniques for improving the size and quality of training datasets for Deep Learning models. It is widely applied as an effective methodology to improve generalization and achieves improvements in many computer vision and natural language processing tasks \cite{Zhang2018mixupBE, Sennrich2016ImprovingNM}. To generate synthetic data and improve the student's performance, we apply WordNet substitution and back translation \cite{ma2019nlpaug} to every distinct sentence in NLI dataset. After data augmentation, the training data size is boosted from 1 million to 3 millions sentences.

\section{Results}
\subsection{Results of STS Tasks}
Table \ref{table:main} presents the results of our model comparing with current state-of-the-art sentence embedding models on STS tasks. Our \modelname{}-MiniLM can achieve 97.7\% of Spearman's correlation performance and 7 times higher speed with only 6.5\% of parameters compared with the best performance model SimCSE-RoBERTa\textsubscript{large}. We also observe that our \modelname{}-TinyBERT and \modelname{}-MiniLM models outperform SimCSE-TinyBERT and SimCSE-MiniLM, which are directly fine-tuned on the same training data and loss function as SimCSE-RoBERTa\textsubscript{large}.
Besides, our results show that our model can significantly improve the results with 2.7-4.5\% absolute gain compared with projection or whitening for dimension reduction.

\paragraph{Impact of Data Augmentation and Final Dimension}  Results in Table \ref{table:data} show that models with augmented data can raise the performance by 1-2 points compared with ones without augmented data. For example, \modelname{}-MiniLM-128 achieves an average Spearman's correlation of 81.20 with data augmentation, compared to 79.48 without data augmentation. We find that different projected layer dimensions achieve similar performances. However, small dimension has slightly better results than large ones.

\begin{table}[tbp] 
\small
\centering
\begin{tabular}{cccc}
\Xhline{2\arrayrulewidth}
Model   & Dim & STS-B & Avg.  \\
\hline\hline
\multirow{3}*{\modelname{}-MiniLM} & 128 & 83.91 & 81.20 \\
& 256  & 83.95  & 81.05 \\
& 384 & 83.44 & 80.91 \\
\hline
\multirow{3}*{\modelname{}-MiniLM-wo-Aug} & 128      & 82.33 & 79.48 \\
& 256      & 82.55 & 79.57 \\
& 384     & 82.04 & 79.15 \\

\hline 
\multirow{3}*{\modelname{}-TinyBERT} & 128 & 83.41 & 80.99 \\
& 256 & 83.19 & 80.81 \\
& 312 & 83.11 & 80.72 \\
\hline
\multirow{3}*{\modelname{}-TinyBERT-wo-Aug} & 128     & 81.88 & 79.64 \\
& 256     & 81.67 & 79.47 \\
& 312     & 81.50 & 79.27 \\

\Xhline{2\arrayrulewidth}
\end{tabular}
\caption{Effect of data augmentation and different dimensions (STS-B and Avg. in STS tasks, wo: without, \modelname{} Teacher: SimCSE-MPNet)}
\label{table:data}
\end{table}

\subsection{Results of SR Tasks}
From Table \ref{table:FAISS}, we demonstrate that the embedding dimension plays a vital role in the performance of semantic retrieval tasks. 
Our \modelname{} model with different dimensions can achieve comparable MRR performance while the retrieval speed and memory usage increase significantly when dimension goes up. Compared with SimCSE-MPNet, which outputs a 768 dimensional vector, our model with 128 dimensions can achieve competitive MRR performance while reducing the retrieval time by 8.2$\times$ and memory usage by 8.0$\times$.

\begin{table}[tbp] 
\small
\centering
\begin{tabular}{cccc}
\Xhline{2\arrayrulewidth}
Model   & MRR   & Time  & Mem   \\
\hline\hline
\modelname{}-TinyBERT-128 & 0.613 & 63.1  & 42.61 \\
\modelname{}-TinyBERT-256 & 0.616 & 130.4 & 85.22 \\
\modelname{}-TinyBERT-312 & 0.615 & 165.4 & 103.86 \\
\hline
\modelname{}-MiniLM-128 & 0.610 & 68.6  & 42.61\\
\modelname{}-MiniLM-256 & 0.615 & 132.1 & 85.22\\
\modelname{}-MiniLM-384 & 0.612 & 194.4 & 127.83\\
\hline 
SimCSE-MPNet-768 & 0.671 & 385.8 & 255.66 \\
SimCSE-RoBERTa\textsubscript{large}-1024 & 0.670 & 518.0 & 340.88\\
\Xhline{2\arrayrulewidth}
\end{tabular}
\caption{Semantic retrieval results on Quora dataset. (MRR@10: retrieval quality, Time: retrieval efficiency, Mem: memory consumption)}
\label{table:FAISS}
\end{table}

\section{Conclusion and Discussion}
In this paper, we propose an effective method to compress sentence representation using homomorphic projective distillation. We demonstrated that this approach successfully enables small language models to achieve competitive high-quality sentence representations compared with large ones while keeping a small embedding size to optimize the memory storage and retrieval latency in downstream tasks.

Our results show that knowledge distillation with augmented data improves the student model's capability to cover and understand more complex sentence variances. The learned projection layer with contrastive loss for sentence embedding can outperform other dimension reduction methods. We also try adding whitening transformation on \modelname{}'s output and the performance is slightly dropped (Appendix \ref{sec:appendix_b}). Since we find that smaller dimensions can have slightly better results than larger ones, we will check over the optimal projected layer size to enhance the isotropy of sentence representation distribution for semantic similarity tasks in our future work.

\section*{Acknowledgements}
We would like to thank the anonymous reviewers for their thoughtful comments. We would also like to thank Yang Gao for valuable feedback on an early version of this
work.
\bibliography{anthology,custom}

\begin{thebibliography}{39}
\expandafter\ifx\csname natexlab\endcsname\relax\def\natexlab#1{#1}\fi

\bibitem[{Abdi and Williams(2010)}]{Abdi2010PrincipalCA}
Herv{\'e} Abdi and Lynne~J. Williams. 2010.
\newblock Principal component analysis.
\newblock \emph{Wiley Interdisciplinary Reviews: Computational Statistics},
  2:433--459.

\bibitem[{Agirre et~al.(2015)Agirre, Banea, Cardie, Cer, Diab, Gonzalez-Agirre,
  Guo, Lopez-Gazpio, Maritxalar, Mihalcea, Rigau, Uria, and
  Wiebe}]{Agirre2015SemEval2015T2}
Eneko Agirre, Carmen Banea, Claire Cardie, Daniel~Matthew Cer, Mona~T. Diab,
  Aitor Gonzalez-Agirre, Weiwei Guo, I{\~n}igo Lopez-Gazpio, Montse Maritxalar,
  Rada Mihalcea, German Rigau, Larraitz Uria, and Janyce Wiebe. 2015.
\newblock Semeval-2015 task 2: Semantic textual similarity, english, spanish
  and pilot on interpretability.
\newblock In \emph{SemEval@NAACL-HLT}.

\bibitem[{Agirre et~al.(2014)Agirre, Banea, Cardie, Cer, Diab, Gonzalez-Agirre,
  Guo, Mihalcea, Rigau, and Wiebe}]{Agirre2014SemEval2014T1}
Eneko Agirre, Carmen Banea, Claire Cardie, Daniel~Matthew Cer, Mona~T. Diab,
  Aitor Gonzalez-Agirre, Weiwei Guo, Rada Mihalcea, German Rigau, and Janyce
  Wiebe. 2014.
\newblock Semeval-2014 task 10: Multilingual semantic textual similarity.
\newblock In \emph{*SEMEVAL}.

\bibitem[{Agirre et~al.(2016)Agirre, Banea, Cer, Diab, Gonzalez-Agirre,
  Mihalcea, Rigau, and Wiebe}]{Agirre2016SemEval2016T1}
Eneko Agirre, Carmen Banea, Daniel~Matthew Cer, Mona~T. Diab, Aitor
  Gonzalez-Agirre, Rada Mihalcea, German Rigau, and Janyce Wiebe. 2016.
\newblock Semeval-2016 task 1: Semantic textual similarity, monolingual and
  cross-lingual evaluation.
\newblock In \emph{*SEMEVAL}.

\bibitem[{Agirre et~al.(2012)Agirre, Cer, Diab, and
  Gonzalez-Agirre}]{Agirre2012SemEval2012T6}
Eneko Agirre, Daniel~Matthew Cer, Mona~T. Diab, and Aitor Gonzalez-Agirre.
  2012.
\newblock Semeval-2012 task 6: A pilot on semantic textual similarity.
\newblock In \emph{*SEMEVAL}.

\bibitem[{Agirre et~al.(2013)Agirre, Cer, Diab, Gonzalez-Agirre, and
  Guo}]{Agirre2013SEM2S}
Eneko Agirre, Daniel~Matthew Cer, Mona~T. Diab, Aitor Gonzalez-Agirre, and
  Weiwei Guo. 2013.
\newblock *sem 2013 shared task: Semantic textual similarity.
\newblock In \emph{*SEMEVAL}.

\bibitem[{Bowman et~al.(2015)Bowman, Angeli, Potts, and
  Manning}]{Bowman2015ALA}
Samuel~R. Bowman, Gabor Angeli, Christopher Potts, and Christopher~D. Manning.
  2015.
\newblock A large annotated corpus for learning natural language inference.
\newblock In \emph{EMNLP}.

\bibitem[{Cer et~al.(2017)Cer, Diab, Agirre, Lopez-Gazpio, and
  Specia}]{Cer2017SemEval2017T1}
Daniel~Matthew Cer, Mona~T. Diab, Eneko Agirre, I{\~n}igo Lopez-Gazpio, and
  Lucia Specia. 2017.
\newblock Semeval-2017 task 1: Semantic textual similarity multilingual and
  crosslingual focused evaluation.
\newblock In \emph{SemEval@ACL}.

\bibitem[{Conneau et~al.(2017)Conneau, Kiela, Schwenk, Barrault, and
  Bordes}]{Conneau2017SupervisedLO}
Alexis Conneau, Douwe Kiela, Holger Schwenk, Lo{\"i}c Barrault, and Antoine
  Bordes. 2017.
\newblock Supervised learning of universal sentence representations from
  natural language inference data.
\newblock In \emph{EMNLP}.

\bibitem[{Devlin et~al.(2019)Devlin, Chang, Lee, and
  Toutanova}]{Devlin2019BERTPO}
Jacob Devlin, Ming-Wei Chang, Kenton Lee, and Kristina Toutanova. 2019.
\newblock Bert: Pre-training of deep bidirectional transformers for language
  understanding.
\newblock In \emph{NAACL}.

\bibitem[{Gao et~al.(2021)Gao, Yao, and Chen}]{gao2021simcse}
Tianyu Gao, Xingcheng Yao, and Danqi Chen. 2021.
\newblock {SimCSE}: Simple contrastive learning of sentence embeddings.
\newblock In \emph{Empirical Methods in Natural Language Processing (EMNLP)}.

\bibitem[{Hao et~al.(2019)Hao, Liu, Wu, and Lv}]{Hao2019ExploitingSE}
Yu~Hao, Xien Liu, Ji~Wu, and Ping Lv. 2019.
\newblock Exploiting sentence embedding for medical question answering.
\newblock In \emph{AAAI}.

\bibitem[{Jiao et~al.(2020)Jiao, Yin, Shang, Jiang, Chen, Li, Wang, and
  Liu}]{jiao-etal-2020-tinybert}
Xiaoqi Jiao, Yichun Yin, Lifeng Shang, Xin Jiang, Xiao Chen, Linlin Li, Fang
  Wang, and Qun Liu. 2020.
\newblock \href {https://doi.org/10.18653/v1/2020.findings-emnlp.372}
  {{T}iny{BERT}: Distilling {BERT} for natural language understanding}.
\newblock In \emph{Findings of the Association for Computational Linguistics:
  EMNLP 2020}, pages 4163--4174, Online. Association for Computational
  Linguistics.

\bibitem[{Johnson et~al.(2017)Johnson, Douze, and J{\'e}gou}]{JDH17}
Jeff Johnson, Matthijs Douze, and Herv{\'e} J{\'e}gou. 2017.
\newblock Billion-scale similarity search with gpus.
\newblock \emph{arXiv preprint arXiv:1702.08734}.

\bibitem[{Khrulkov et~al.(2020)Khrulkov, Mirvakhabova, Ustinova, Oseledets, and
  Lempitsky}]{Khrulkov2020HyperbolicIE}
Valentin Khrulkov, Leyla Mirvakhabova, E.~Ustinova, I.~Oseledets, and Victor~S.
  Lempitsky. 2020.
\newblock Hyperbolic image embeddings.
\newblock \emph{2020 IEEE/CVF Conference on Computer Vision and Pattern
  Recognition (CVPR)}, pages 6417--6427.

\bibitem[{Kiros et~al.(2015)Kiros, Zhu, Salakhutdinov, Zemel, Urtasun,
  Torralba, and Fidler}]{Kiros2015SkipThoughtV}
Ryan Kiros, Yukun Zhu, Ruslan Salakhutdinov, Richard~S. Zemel, Raquel Urtasun,
  Antonio Torralba, and Sanja Fidler. 2015.
\newblock Skip-thought vectors.
\newblock In \emph{NIPS}.

\bibitem[{Liu et~al.(2019)Liu, Ott, Goyal, Du, Joshi, Chen, Levy, Lewis,
  Zettlemoyer, and Stoyanov}]{Liu2019RoBERTaAR}
Yinhan Liu, Myle Ott, Naman Goyal, Jingfei Du, Mandar Joshi, Danqi Chen, Omer
  Levy, Mike Lewis, Luke Zettlemoyer, and Veselin Stoyanov. 2019.
\newblock Roberta: A robustly optimized bert pretraining approach.
\newblock \emph{ArXiv}, abs/1907.11692.

\bibitem[{Logeswaran and Lee(2018)}]{Logeswaran2018AnEF}
Lajanugen Logeswaran and Honglak Lee. 2018.
\newblock An efficient framework for learning sentence representations.
\newblock \emph{ArXiv}, abs/1803.02893.

\bibitem[{Loshchilov and Hutter(2019)}]{Loshchilov2019DecoupledWD}
Ilya Loshchilov and Frank Hutter. 2019.
\newblock Decoupled weight decay regularization.
\newblock In \emph{ICLR}.

\bibitem[{Ma(2019)}]{ma2019nlpaug}
Edward Ma. 2019.
\newblock Nlp augmentation.
\newblock https://github.com/makcedward/nlpaug.

\bibitem[{Marelli et~al.(2014)Marelli, Menini, Baroni, Bentivogli, Bernardi,
  and Zamparelli}]{Marelli2014ASC}
Marco Marelli, Stefano Menini, Marco Baroni, Luisa Bentivogli, Raffaella
  Bernardi, and Roberto Zamparelli. 2014.
\newblock A sick cure for the evaluation of compositional distributional
  semantic models.
\newblock In \emph{LREC}.

\bibitem[{Mikolov et~al.(2013)Mikolov, Chen, Corrado, and
  Dean}]{Mikolov2013EfficientEO}
Tomas Mikolov, Kai Chen, Gregory~S. Corrado, and Jeffrey Dean. 2013.
\newblock Efficient estimation of word representations in vector space.
\newblock In \emph{ICLR}.

\bibitem[{Pagliardini et~al.(2018)Pagliardini, Gupta, and
  Jaggi}]{Pagliardini2018UnsupervisedLO}
Matteo Pagliardini, Prakhar Gupta, and Martin Jaggi. 2018.
\newblock Unsupervised learning of sentence embeddings using compositional
  n-gram features.
\newblock In \emph{NAACL}.

\bibitem[{Palangi et~al.(2016)Palangi, Deng, Shen, Gao, He, Chen, Song, and
  Ward}]{Palangi2016DeepSE}
Hamid Palangi, Li~Deng, Yelong Shen, Jianfeng Gao, Xiaodong He, Jianshu Chen,
  Xinying Song, and Rabab~Kreidieh Ward. 2016.
\newblock Deep sentence embedding using long short-term memory networks:
  Analysis and application to information retrieval.
\newblock \emph{IEEE/ACM Transactions on Audio, Speech, and Language
  Processing}, 24:694--707.

\bibitem[{Raunak and Gupta(2019)}]{Raunak2019EffectiveDR}
Vikas Raunak and Vivek Gupta. 2019.
\newblock Effective dimensionality reduction for word embeddings.
\newblock In \emph{RepL4NLP@ACL}.

\bibitem[{Reimers(2019)}]{SentenceTransformer}
Nils Reimers. 2019.
\newblock Ukplab sentence-transformers.
\newblock https://www.sbert.net/docs/pretrained\_models.html.

\bibitem[{Reimers and Gurevych(2019)}]{reimers-2019-sentence-bert}
Nils Reimers and Iryna Gurevych. 2019.
\newblock \href {https://arxiv.org/abs/1908.10084} {Sentence-bert: Sentence
  embeddings using siamese bert-networks}.
\newblock In \emph{Proceedings of the 2019 Conference on Empirical Methods in
  Natural Language Processing}. Association for Computational Linguistics.

\bibitem[{Sanh et~al.(2019)Sanh, Debut, Chaumond, and
  Wolf}]{Sanh2019DistilBERTAD}
Victor Sanh, Lysandre Debut, Julien Chaumond, and Thomas Wolf. 2019.
\newblock Distilbert, a distilled version of bert: smaller, faster, cheaper and
  lighter.
\newblock \emph{ArXiv}, abs/1910.01108.

\bibitem[{Sennrich et~al.(2016)Sennrich, Haddow, and
  Birch}]{Sennrich2016ImprovingNM}
Rico Sennrich, Barry Haddow, and Alexandra Birch. 2016.
\newblock Improving neural machine translation models with monolingual data.
\newblock \emph{ArXiv}, abs/1511.06709.

\bibitem[{Shi et~al.(2018)Shi, Sun, and Hu}]{Shi2018UnderstandingAI}
Haoyue Shi, Yuqi Sun, and Junfeng Hu. 2018.
\newblock Understanding and improving multi-sense word embeddings via extended
  robust principal component analysis.
\newblock \emph{ArXiv}, abs/1803.01255.

\bibitem[{Song et~al.(2020)Song, Tan, Qin, Lu, and Liu}]{Song2020MPNetMA}
Kaitao Song, Xu~Tan, Tao Qin, Jianfeng Lu, and Tie-Yan Liu. 2020.
\newblock Mpnet: Masked and permuted pre-training for language understanding.
\newblock \emph{ArXiv}, abs/2004.09297.

\bibitem[{Su et~al.(2021)Su, Cao, Liu, and Ou}]{Su2021WhiteningSR}
Jianlin Su, Jiarun Cao, Weijie Liu, and Yangyiwen Ou. 2021.
\newblock Whitening sentence representations for better semantics and faster
  retrieval.
\newblock \emph{ArXiv}, abs/2103.15316.

\bibitem[{Wang and Kuo(2020)}]{Wang2020SBERTWKAS}
Bin Wang and C.-C.~Jay Kuo. 2020.
\newblock Sbert-wk: A sentence embedding method by dissecting bert-based word
  models.
\newblock \emph{IEEE/ACM Transactions on Audio, Speech, and Language
  Processing}, 28:2146--2157.

\bibitem[{Wang et~al.(2020)Wang, Wei, Dong, Bao, Yang, and
  Zhou}]{Wang2020MiniLMDS}
Wenhui Wang, Furu Wei, Li~Dong, Hangbo Bao, Nan Yang, and Ming Zhou. 2020.
\newblock Minilm: Deep self-attention distillation for task-agnostic
  compression of pre-trained transformers.
\newblock \emph{ArXiv}, abs/2002.10957.

\bibitem[{Williams et~al.(2018)Williams, Nangia, and Bowman}]{Williams2018ABC}
Adina Williams, Nikita Nangia, and Samuel~R. Bowman. 2018.
\newblock A broad-coverage challenge corpus for sentence understanding through
  inference.
\newblock In \emph{NAACL}.

\bibitem[{Wolf et~al.(2020)Wolf, Debut, Sanh, Chaumond, Delangue, Moi, Cistac,
  Rault, Louf, Funtowicz, and Brew}]{Wolf2020TransformersSN}
Thomas Wolf, Lysandre Debut, Victor Sanh, Julien Chaumond, Clement Delangue,
  Anthony Moi, Pierric Cistac, Tim Rault, R{\'e}mi Louf, Morgan Funtowicz, and
  Jamie Brew. 2020.
\newblock Transformers: State-of-the-art natural language processing.
\newblock In \emph{EMNLP}.

\bibitem[{Yang et~al.(2020)Yang, Wang, Zhou, Zhao, Yu, Zhang, and
  Li}]{Yang2020TowardsMT}
Jiacheng Yang, Mingxuan Wang, Hao Zhou, Chengqi Zhao, Yong Yu, Weinan Zhang,
  and Lei Li. 2020.
\newblock Towards making the most of bert in neural machine translation.
\newblock In \emph{AAAI}.

\bibitem[{Zhang et~al.(2018)Zhang, Ciss{\'e}, Dauphin, and
  Lopez-Paz}]{Zhang2018mixupBE}
Hongyi Zhang, Moustapha Ciss{\'e}, Yann Dauphin, and David Lopez-Paz. 2018.
\newblock mixup: Beyond empirical risk minimization.
\newblock \emph{ArXiv}, abs/1710.09412.

\bibitem[{Zhao et~al.(2020)Zhao, Xue, Yu, Li, and Yang}]{zhao2020multi}
Xuandong Zhao, Jinbao Xue, Jin Yu, Xi~Li, and Hongxia Yang. 2020.
\newblock A multi-semantic metapath model for large scale heterogeneous network
  representation learning.
\newblock \emph{arXiv preprint arXiv:2007.11380}.

\end{thebibliography}
\bibliographystyle{acl_natbib}

\appendix

\begin{table*}[thbp] 
\small
\centering
\begin{tabular}{cccccccc|c}
\Xhline{2\arrayrulewidth} 
Model & STS12 & STS13 & STS14 & STS15 & STS16 & STS-B & SICK-R & Avg.\\ 
\hline\hline
\modelname{}-MiniLM-H128& 	74.25& 	84.43& 	80.33& 	85.75& 	80.68& 	83.91& 	79.06& 	81.20\\
\modelname{}-MiniLM-H256& 	73.95& 	84.21& 	80.04& 	86.08& 	81.11& 	83.95& 	78.89& 	81.05\\
\modelname{}-MiniLM-H384& 	73.63& 	83.91& 	79.71& 	85.90& 	80.88& 	83.44& 	78.88& 	80.91\\
\hline
\modelname{}-MiniLM-H128-wo-Aug& 71.39 & 82.45 &	78.24& 	84.65& 	78.85& 	82.33& 	78.42& 	79.48\\
\modelname{}-MiniLM-H256-wo-Aug& 	71.36& 	82.65& 	78.20& 	84.65& 	79.21& 	82.55& 	78.36& 	79.57\\
\modelname{}-MiniLM-H384-wo-Aug& 	70.94& 	82.06& 	77.60& 	84.41& 	78.70& 	82.04& 	78.31& 	79.15\\
\hline
\modelname{}-TinyBERT-H128& 	74.2& 	84.49& 	79.95& 	85.79& 	80.07& 	83.41& 	78.99& 	80.99\\
\modelname{}-TinyBERT-H256& 	74.06& 	84.14& 	79.7& 	85.93& 	80.03& 	83.19& 	78.60& 	80.81\\
\modelname{}-TinyBERT-H312& 	73.97& 	84.14& 	79.61& 	85.65& 	79.79& 	83.11& 	78.74& 	80.72\\
\hline
\modelname{}-TinyBERT-H128-wo-Aug& 	73.29& 	82.51& 	78.36& 	84.61& 	78.45& 	81.88& 	78.39& 	79.64\\
\modelname{}-TinyBERT-H256-wo-Aug& 	73.00& 	82.25& 	78.36& 	84.74& 	78.10& 	81.67& 	78.20& 	79.47\\
\modelname{}-TinyBERT-H312-wo-Aug& 	72.85& 	82.20& 	77.90& 	84.35& 	77.83& 	81.50& 	78.23& 	79.27\\
\hline
\modelname{}-MiniLM-H384-whiten-128 & 73.73 & 84.10 & 79.47 & 85.23 & 79.32 & 82.69 & 78.74 & 80.47\\
\modelname{}-MiniLM-H384-whiten-256 & 73.98 & 84.15 & 79.61 & 85.63 & 79.78 & 83.09 & 78.73 & 80.71\\
\modelname{}-TinyBERT-H312-whiten-128 & 73.91 & 84.08 & 79.52 & 85.32 & 79.45 & 82.81 & 78.78 & 80.55\\
\modelname{}-TinyBERT-H312-whiten-256 & 74.00 & 84.15 & 79.62 & 85.64 & 79.77 & 83.09 & 78.74 & 80.72\\
\Xhline{2\arrayrulewidth}
\end{tabular}
\caption{Sentence embedding performance on STS tasks (Spearman's correlation $\rho \times 100$).}
\vspace{-4mm}
\label{table:app}
\end{table*}

\section{Training Details}
\label{sec:appendix_a}
 We elaborate on how we obtain different baselines for comparisons in Table \ref{table:main}.
\begin{itemize}
    \item For SBERT\textsubscript{base} and SRoBERTa\textsubscript{large}, we 
report the results from \citet{reimers-2019-sentence-bert} and test their speed based on released models.
    \item For SimCSE-RoBERTa\textsubscript{large}, we directly load the pre-trained models from Huggingface’s repository \cite{Wolf2020TransformersSN} "princeton-nlp/sup-simcse-roberta-large".
    \item For SimCSE-MPNet, we utilize a well fine-tuned sentence embedding model using contrastive loss trained on NLI dataset from Huggingface’s repository "sentence-transformers/nli-mpnet-base-v2".
    \item For SimCSE-MiniLM, we use the MiniLM with 6 layers, 384-hidden size and 6 self-attention heads as the backbone network. We then fine-tune it following the contrastive loss for 3 epochs with a batch size of 256. The optimizer we use is AdamW \cite{Loshchilov2019DecoupledWD} and the learning rate is set as 1e-3.
    \item For SimCSE-TinyBERT, we use the TinyBERT with 4 layers, 312-hidden size and 12 self-attention heads. The other training settings are the same as SimCSE-MiniLM.
    \item For Projection-128, we add a linear layer to the 
    language model MiniLM/TinyBERT. The linear layer projects the original embedding from 384/312 dimension to 128 dimension. We train the model using the same contrastive loss and configuration as those of SimCSE-MiniLM/SimCSE-TinyBERT.
    \item For Whitening-128, we implement our own version of whitening operation \cite{Su2021WhiteningSR}. It is directly applied on SimCSE-MiniLM/SimCSE-TinyBERT as a dimension reduction technique. Note that whitening is a post-processing method, which is different from \modelname{}.
    \item For \modelname{}-MiniLM and \modelname{}-TinyBERT, the models are trained for 3 epochs with a batch size of 256 and a learning rate of 1e-4. We keep the best checkpoint during training by evaluating the model on STS-B test sets.
\end{itemize}

\section{More Results on STS Tasks}
\label{sec:appendix_b}
We report the full set of results for data augmentation and different dimensions on STS tasks in Table \ref{table:app} (Teacher model: SimCSE-MPNet). We also test a variation: adding whitening after the projected distillation. Results show that adding whitening after our \modelname{} output slightly decreases the performance.


\label{sec:appendix_d}


\end{document}